\documentclass[12pt]{article}
\usepackage{amsmath}
\usepackage{times}
\usepackage{graphicx}
\usepackage{color}
\usepackage{multirow}
\usepackage[authoryear]{natbib}
\usepackage{rotating}
\usepackage{bbm}
\usepackage{latexsym}
\usepackage{verbatim}
\usepackage{enumerate}
\usepackage{subfig}

\newcommand{\captionfonts}{\normalsize}

\makeatletter  
\long\def\@makecaption#1#2{%
  \vskip\abovecaptionskip
  \sbox\@tempboxa{{\captionfonts #1: #2}}%
  \ifdim \wd\@tempboxa >\hsize
    {\captionfonts #1: #2\par}
  \else
    \hbox to\hsize{\hfil\box\@tempboxa\hfil}%
  \fi
  \vskip\belowcaptionskip}
\makeatother   

\begin{document}
\hspace{13.9cm}1

\ \vspace{20mm}\\

{\LARGE Supervised Learning in Multilayer Spiking Neural Networks}

\ \\
{\bf \large {Ioana Sporea$^{\displaystyle 1}$} and Andr\'e Gr\"uning$^{\displaystyle 1}$}\\
{$^{\displaystyle 1}$Department of Computing, University of Surrey.}\\
%


\thispagestyle{empty}
\markboth{}{}
\ \vspace{-0mm}\\
%
\begin{center} {\bf Abstract} \end{center}
The current article introduces a supervised learning algorithm for multilayer spiking neural networks. The algorithm presented here overcomes some limitations of existing learning algorithms as it can be applied to neurons firing multiple spikes and it can in principle be applied to any linearisable neuron model. The algorithm is applied successfully to various benchmarks, such as the XOR problem and the Iris data set, as well as complex classifications problems. The simulations also show the flexibility of this supervised learning algorithm which permits different encodings of the spike timing patterns, including precise spike trains encoding.

\section{Introduction}
\label{sec:1}
\par
Traditional rate coded artificial neural networks represent an analog variable through the firing rate of the biological neuron. That is, the output of a computational unit is a representation of the firing rate of the biological neuron. In order to increase the computational power of the network the neurons are structured in successive layers of computational units. Such systems are trained to recognise the input patterns by searching for a set of suitable connections weights. Learning rules based on gradient decent, such as backpropagation \citep{Rumelhart86}, have led sigmoidal neural networks (networks that use the sigmoid as the activation function) to be one of the most powerful and flexible computational models. 

\par
However, experimental evidence suggests that neural systems use the exact time of single action potentials to encode information \citep{Thorpe89, Johansson04}. In \cite{Thorpe89} it is argued that because of the speed of processing visual information and the anatomical structure of the visual system, processing has to be done on the basis of single spikes. In \cite{Johansson04} it is shown that the relative timing of the first spike contains important information about tactile stimuli. Further evidence suggests that the precise temporal firing pattern of groups of neurons conveys relevant sensory information \citep{Wehr96, Neuenschwander96, deCharms96}.

\par
These findings have led to a new way of simulating neural networks based on temporal encoding with single spikes \citep{Maass97a}. Investigations of the computational power of spiking neurons have illustrated that realistic mathematical models of neurons can arbitrarily approximate any continuous function, and furthermore, it has been demonstrated that networks of spiking neurons are computationally more powerful than sigmoidal neurons \citep{Maass97b}. Because of the nature of spiking neuron communication, these are also suited for VSLI implementation with significant speed advantages \citep{Elias02}.

\par
In this paper, we present a new learning algorithm for feed-forward spiking neural networks with multiple layers. The learning rule extends the ReSuMe algorithm \citep{Ponulak10} to multiple layers using backpropagation of the network error. The weights are updated according to STDP and anti-STDP processes and unlike SpikeProp \citep{Bohte02} can be applied to neurons firing multiple spikes in all layers. The multilayer ReSuMe is analogue of the backpropagation learning algorithm for rate neurons, while making use of spiking neurons. To the best of our knowledge this is the first learning algorithm for spiking neuron networks with hidden layers where multiple spikes are considered in all layers, with precise spike time encoding can be read for both inputs and outputs.

\par
The rest of the article is organised as follows: in the following section some of the existing supervised learning algorithms for spiking neurons are discussed. Section \ref{sec:3} contains a description of the generalised spiking neuron model and the derivation of the learning rule based on this neuron model for a feed-forward network with a hidden layer. In section \ref{sec:4} the weight modifications are analysed for a simplified network with a single output neuron. In section \ref{sec:5} the flexibility and power of the feed-forward spiking neural networks trained with multilayer ReSuMe are showcased by non-linear problems and classifications tasks. The spiking neural network is trained with spike timing patterns distributed over timescales in the range of tens to hundreds of milliseconds, comparable to the span of sensory and motor processing \citep{Mauk04}. A discussion of the learning algorithm and the results are presented in section \ref{sec:6}. The article concludes with a summary of the article.

\section{Background}
\label{sec:2}
\par
Whereas experimental studies have shown that supervised learning is present in the brain, especially in sensorimotor networks and sensory systems \citep{Knudsen94, Knudsen02}, there are no definite conclusions regarding the means through which biological neurons learn. Several learning algorithms have been proposed to explore how spiking neurons may respond to given instructions. 
\par
One such algorithm, the tempotron learning rule, has been introduced by \cite{Gutig06}, where neurons learn to discriminate between spatiotemporal sequences of spike patterns. Although the learning rule uses a gradient-descent approach, it can only be applied to single-layered networks. The algorithm is used to train leaky integrate-and-fire neurons to distinguish between two classes of patterns by firing at least one action potential or by remaining quiescent. While the spiking neural network is able to successfully classify the spike-timing patterns, the neurons do not learn to respond with precise spike-timing patterns. 

\par
Another gradient descent based learning rule is the SpikeProp algorithm \citep{Bohte02} and its extensions \citep{Schrauwen04, Xin01, Booij05, Tino05}. The algorithm is applied to feed-forward networks of neurons firing a single spike and is minimizing the time difference between the target spike and the actual output spike time. The learning algorithm uses spiking neurons modelled by the Spike Response Model \citep{Gerstner01}, and the derivations of the learning rule are based on the explicit dynamics of the neuron model. Although \cite{Booij05} have extended the algorithm to allow neurons to fire multiple spikes in the input and hidden layers, subsequent spikes in the output layer are still ignored because the network error is represented by the time difference of the first spike of the target and output neurons. 

\par
Some supervised learning algorithms are based on Hebb's postulate - "\emph{cells that fire together, wire together}" \citep{Hebb49} - \citep{Ruf97, Legenstein05, Ponulak10}. ReSuMe \citep{Ponulak10} is making use of both Hebbian learning and gradient descent techniques. As the weight modifications are based only on the input and output spike trains and do not make any explicit assumptions about the neural or synaptic dependencies, the algorithm can be applied to various neuron models. However, the algorithm can only be applied to a single layer of neurons or used to train readouts for reservoir networks. 

\par
The ReSuMe algorithm has also been applied to neural networks with a hidden layer, where weights of downstream neurons are subject to multiplicative scaling \citep{Gruning11}. The simulations show that networks with one hidden layer can perform non-linear logical operations, while networks without hidden layers cannot. The ReSuMe algorithm has also been used to train the output layer in a layer-feedforward network in \cite{glackin:11} and \cite{wade:10} where the  hidden layer acted as a frequency filter. However, input and target outputs here consisted of fixed-rate spike trains. 

\par
Finally, there is a linear programming approach to estimating weights (and delays) in recurrent spiking networks based on LIF neurons \citep{rostro:10} which successfully attempts to reconstruct weights from a spike raster and initial conditions such as a (fixed) input spike train and initial states of membrane potentials. 

\par
The present paper introduces a new supervised learning algorithm that combines the quality of SpikeProp, spanning to multiple layers \citep{Bohte02}, with the flexibility of ReSuMe, which can be used with multiple spikes and different neuron models \citep{Ponulak10}.

\section{Learning algorithm}
\label{sec:3}
\par
In this section the new learning algorithm for feed-forward multilayer spiking neural networks is described. The learning rule is derived for networks with only one hidden layer, as the algorithm can be extended to networks with more hidden layers similarly.
 
\subsection{Neuron model}
\label{sec:3.1}
\par
The input and output signals of spiking neurons are represented by the timing of spikes. A spike train is defined as a sequence of impulses fired by a particular neuron at times $t^f$. Spike  trains are formalised by a sum of Dirac delta functions \citep{Gerstner02}:

\begin{equation}
S(t)=\sum_f \delta(t-t^f)
\label{eq:1}
\end{equation}
\par
In order to analyse the relation between the input and output spike trains, we use the linear Poisson neuron model \citep{Gutig03, Kempter01}. Its instantaneous firing rate $R(t)$ is formally defined as the expectation of the spike train, averaged over an infinite number of trials. An estimate of the instantaneous firing rate can be obtained by averaging over a finite number of trials \citep{Heeger01}:  
\begin{equation}
R(t)=<S(t)>=\frac{1}{M}\sum_{j=1}^MS_j(t)
\label{eq:3}
\end{equation}
where $M$ is the number of trials and $S_j(t)$ is the concrete spike train for each trial. In the linear Poisson model the spiking activity of the postsynaptic neuron $o$ is defined by the instantaneous rate function: 

\begin{equation}
R_o(t) = \frac{1}{n}\sum_{h\in H} w_{oh}R_h(t)
\label{eq:2}
\end{equation}
where $n$ is the number of presynaptic neurons $h$. The weights $w_{oh}$ represent the strength of the connection between the presynaptic neurons $h$ and postsynaptic neuron $o$. The instantaneous firing rate $R(t)$ will be used for the derivation of the learning algorithm due to its smoothness and subsequently be replaced by its discontinuous estimate, the spike train $S(t)$.

\subsection{Backpropagation of the network error}
\label{sec:3.2}
\par
The learning algorithm is derived for a fully connected feed-forward network with one hidden layer. The input layer $I$ is only setting the input patterns without performing any computation on the patterns. The hidden and output layers are labelled $H$ and $O$ respectively. All neurons in one layer are connected to all neurons in the subsequent layer.

\par
The instantaneous network error is formally defined in terms of the difference between the actual instantaneous firing rate $R_o^a(t)$ and the target instantaneous firing rate $R_o^d(t)$ for all output neurons:
\begin{equation}
E(t)=E(R_o^a(t))=\frac{1}{2}\sum_{o\in O}\left(R_o^a(t)-R_o^d(t)\right)^2
\label{eq:4}
\end{equation}
In order to minimise the network error, the weights are modified using a process of gradient descent:
\begin{equation}
\Delta w_{oh}(t)=-\eta\frac{\partial E(R_o^a(t))}{\partial w_{oh}}
\label{eq:5}
\end{equation}
where $\eta$ is the learning rate and $w_{oh}$ represents the weight between the output neuron~$o$ and hidden neuron~$h$. $\Delta w_{oh}(t)$ is the weight change contribution due to the error $E(t)$ at time $t$, and the total weight change is $\Delta w=\int\Delta w(t)dt$ over the duration of the spike train. This is analogue to  the starting point of standard backpropagation for rate neurons in discrete time. For simplicity, the learning rate will be considered $\eta=1$ and will be suppressed in all following equations, as the step length of each learning iteration will be given by other learning parameters to be defined later on. Also, in the following derivatives are understood in a functional sense.

\subsubsection*{Weight modifications for the output neurons}

In this section we re-derive the weight-update formulate for the ReSuMe learning algorithm and connect with gradient-descent learning for linear Poisson-neurons. We will need this derivation as a first step to a derive our extension of ReSuMe to subsequent layers in the next subsection. However, this derivation is also instructive in its own right as it works out a bit more rigorously than in the original derivation \citep{Ponulak10} how ReSuMe and gradient descent are connected. It also makes Ponulak's statement more explicit that ReSuMe can be applied to any neuron model. This is then the case if the neural model can on an appropriate time scale be approximated well enough with a linear neuron model. 

\par
As the network error is a function of the output spike train, which in turn depends on the weight $w_{oh}$, the derivative of the error function can be expanded using the chain rule as follows:
\begin{equation}
\frac{\partial E(R_o^a(t))}{\partial w_{oh}}=\frac{\partial E(R_o^a(t))}{\partial R_o^a(t)}\frac{\partial R_o^a(t)}{\partial w_{oh}}
\label{eq:6}
\end{equation}
The first term of the right-hand part of equation \eqref{eq:6} can be calculated as:
\begin{equation}
\frac{\partial E(R_o^a(t))}{\partial R_o^a(t)}=R_o^a(t) - R_o^d(t)
\label{eq:7}
\end{equation}
Since the instantaneous rate function is expressed in terms of the weight $w_{oh}$ in \eqref{eq:2}, the second factor of the right-hand side of equation \eqref{eq:6} becomes:
\begin{equation}
\frac{\partial R_o(t)}{\partial w_{oh}} = \frac{1}{n_h}R_h(t)
\label{eq:8}
\end{equation}
where $n_h$ is the number of hidden neurons. By combining equations \eqref{eq:5} -- \eqref{eq:8}, the formula for weight modifications to the output neurons becomes:
\begin{equation}
\Delta w_{oh}(t)=-\frac{1}{n_h}\left[R_o^a(t)-R_o^d(t)\right]R_h(t)
\label{eq:9}
\end{equation}
For convenience we define the backpropagated error $\delta_o(t)$ for the output neuron $o$:
\begin{equation}
\delta_o(t) := \frac{1}{n_h}\left[R_o^d(t)-R_o^a(t)\right]
\end{equation}
hence:
\begin{equation}
\Delta w_{oh}(t)=\delta_o(t)R_h(t)
\end{equation}
This is similar to standard discrete-time backpropagation, however now derived as a functional derivative in continuous time. In the following we will use the best estimation of the unknown instantaneous firing rate $R(t)$ when we only have a single spike train $S(t)$, which is the spike train itself for each of the neurons involved. Thus the weights will be modified according to:
\begin{equation}
\Delta w_{oh}(t)=\frac{1}{n_h}\left[S_o^d(t)-S_o^a(t)\right]S_h(t)
\label{eq:10}
\end{equation}
However, products of Dirac $\delta$ functions are mathematically problematic. Following \cite{Ponulak10} the non-linear product of $S_o^d(t)S_h(t)$ is substituted with a STDP process. In a similar manner, $S_o^a(t)S_h(t)$ is substituted with an anti-STDP process (for details see \cite{Ponulak10}).  
\begin{equation}
\begin{split}
S_o^d(t)S_h(t) \to &  S_h(t)\left[a + \int_0^\infty a^{pre}(s)S_o^d(t-s)ds \right] \\
                  & + S_o^d(t)\left[a + \int_0^\infty a^{post}(s)S_h(t-s)ds \right] 
\end{split}
\end{equation}

\begin{equation}
\begin{split}
S_o^a(t)S_h(t) \to & - S_h(t)\left[a + \int_0^\infty a^{pre}(s)S_o^a(t-s)ds \right] \\
                   & - S_o^a(t)\left[a + \int_0^\infty a^{post}(s)S_h(t-s)ds \right] 
\end{split}
\end{equation}
where $a>0$ is a non-Hebbian term that guarantees the weight changes in the correct direction if the output spike train contains more or less spikes than the target spike train. 

\par
The integration variable $s$ represents the time difference between the actual firing time of the output neuron and the firing time of the hidden neuron $s=(t_o^f-t_h^f)$, and the target firing time and the firing time of the hidden neuron $s=(t_d^f-t_h^f)$ respectively. The kernel $a^{pre}(s)$ gives the weight change if the presynaptic spike (the spike of the hidden neuron occurs) comes after the postsynaptic spike (the spikes of the output and target neurons). The kernel $a^{post}(s)$ gives the weight change if the presynaptic spike before the postsynaptic spike. The kernels $a^{pre}$ and $a^{post}$ define the learning window $W(s)$ \citep{Gerstner02}:
\begin{equation}
W(s)=\begin{cases}
a^{pre}(-s) = -A_{-}\exp(\frac{s}{\tau_-}), & \mbox{if } s\leq 0 \\
a^{post}(s) = +A_{+}\exp(\frac{-s}{\tau_+}), & \mbox{if } s>0 
\end{cases}
\label{eq:15w}
\end{equation}
where $A_+$, $A_->0$ are the amplitudes and $\tau_+$, $\tau_->0$ are the time constants of the learning window. Thus the final learning formula for the weight modifications becomes:
\begin{equation}
\label{eq:12}
\begin{split}
\Delta w_{oh}(t) =& \frac{1}{n_h}S_h(t)\left[\int_0^\infty a^{pre}[S_o^d(t)-S_o^a(t)]ds\right] \\
                  & + \frac{1}{n_h}\left[S_o^d(t)-S_o^a(t)\right]\left[a+\int_0^\infty a^{post}(s)S_h(t-s)ds\right]
\end{split}
\end{equation}

\par
The total weight change is obtained by integrating equation \eqref{eq:12} over time on a time domain that covers all the spikes in the system. This equation is the core of ReSuMe learning algorithm as stated in \citet{Ponulak10}. 

\subsubsection*{Weight modifications for the hidden neurons}
\par
In this section we extend the argument above to weight changes between the input and the hidden layer. The weight modifications for the hidden neurons are calculated in a similar manner in the negative gradient direction:
\begin{equation}
\Delta w_{hi}(t)=-\frac{\partial E(R_o^a(t))}{\partial w_{hi}}
\label{eq:13}
\end{equation}
The derivative of the error is expanded similarly as in equation \eqref{eq:6} (again in the sense of functional derivatives):
\begin{equation}
\frac{\partial E(R_o^a(t))}{\partial w_{hi}}=\frac{\partial E(R_o^a(t))}{\partial R_h(t)}\frac{\partial R_h(t)}{\partial w_{hi}}
\label{eq:14}
\end{equation}
The first factor of the right-hand part of the above equation is expanded for each output neuron using the chain rule:
\begin{equation}
\frac{\partial E(R_o^a(t))}{\partial R_h(t)}=\sum_{o\in O}\frac{\partial E(R_o^a(t))}{\partial R_o^a(t)}\frac{\partial R_o^a(t)}{\partial R_h(t)}
\label{eq:15}
\end{equation}
The second factor of the right-hand side of the above equation is calculated from equation \eqref{eq:2}:
\begin{equation}
\frac{\partial R_o^a(t)}{\partial R_h(t)}=\frac{1}{n_h}w_{oh}
\label{eq:16}
\end{equation}
The derivatives of the error with respect to the output spike train have already been calculated for the weights to the output neurons in equation \eqref{eq:7}. By combining these results:
\begin{equation}
\frac{\partial E(R_o^a(t))}{\partial R_h(t)}=\frac{1}{n_h}\sum_{o\in O}\left[R_o^a(t)-R_o^d(t)\right]w_{oh}
\label{eq:17}
\end{equation}
The second factor of the right-hand part of equation \eqref{eq:14} is calculated as follows using again equation \eqref{eq:2}:
\begin{equation}
\frac{\partial R_h(t)}{\partial w_{hi}}= \frac{1}{n_i}R_i(t)
\label{eq:18}
\end{equation}
where $n_i$ is the number of input neurons. By combining equations \eqref{eq:13} -- \eqref{eq:18}, the formula for the weight modifications to the hidden neurons becomes:
\begin{equation}
\Delta w_{hi}(t)=-\frac{1}{n_hn_i}\sum_{o\in O}\left[R_o^a(t) - R_o^d(t)\right]R_i(t)w_{oh}
\label{eq:19}
\end{equation}
We define the backpropagated error $\delta_h(t)$ for layers other than the output layer:
\begin{equation}
\delta_h(t):=\frac{1}{n_i}\sum_{o\in O}\delta_o(t) w_{oh}
\end{equation}

\par
Just like in standard backpropagation $\delta_o(t)$ are backpropagated errors of the neurons in the preceding layer. By substituting the instantaneous firing rates with the spike trains as estimators, equation \eqref{eq:19} becomes:
\begin{equation}
\Delta w_{hi}(t)=\frac{1}{n_hn_i}\sum_{o\in O}\left[S_o^d(t) - S_o^a(t)\right]S_i(t)w_{oh}
\label{eq:19a}
\end{equation}
We now want to repeat the procedure of replacing the product of two spike trains (involving $delta$-distributions) with a STDP process. We note first that equation \eqref{eq:19a} does not depend any longer on any spikes fired or not fired in the hidden layer. While there are neurobiological plasticity processes that can convey information about a transmitted spike from the effected synapses to lateral or downstream synapses (for an overview see \citealt*{harris:08}), no direct neurobiological basis is known for an STDP process between a synapse and the outgoing spikes of an upstream neuron. Therefore this  substitution is to be seen as a computational analogy, and the weights will be modified according to: 
\begin{equation}
\begin{split}
\Delta w_{hi}(t)= & \frac{1}{n_in_h}S_i(t)\sum_{o\in O}\left[\int_0^\infty a^{pre}[S_o^d(t)-S_o^a(t)]\right]w_{oh} \\
                  & + \frac{1}{n_in_h}\sum_{o\in O}\left[S_o^d(t) - S_o^a(t)\right]\left[a+\int_0^\infty a^{post}(s)S_i(t-s)ds\right]w_{oh}
\end{split}
\label{eq:20}
\end{equation}
\par
The total weight change is again determined by integrating equation \eqref{eq:20} over time. The synaptic weights  between the input and hidden neurons are modified according to STDP processes between the input and target spikes and anti-STDP processes between input and output spikes. 

\subsubsection*{Normalisation}

The normalisation to the number of presynaptic connections of the modifications of the weights to the output neurons ensures that the changes are proportional to the number of weights. Moreover, the learning parameters do not need to change as the network architecture changes (for example, in order to keep the firing rate of postsynaptic neurons constant as the number of presynaptic units changes, the initial weights and weight modifications also must change accordingly). The normalisation to the number of presynaptic and postsynaptic connections of the weight modifications to the hidden neurons ensures that the changes of the connections between the input and hidden layer are usually smaller than the changes of the connections between the hidden and output layer, which keeps the learning process stable.

\subsubsection*{Generalisation}

The algorithm can be generalised in this manner for neural networks with multiple hidden layers. The learning rule could also be generalised for recurrent connections (e.g. using unrolling in time as in backpropagation through time \citep{Rojas96}), however in the present paper we only consider feed-forward connections. This is our extension of ReSuMe to hidden layers following from error minimisation and gradient descent. 

\par
As the learning rule for the weight modifications depends only on the presynaptic and postsynaptic spike trains and the current strength of the connections between the spiking neurons, the algorithm can be applied to various  spiking neuron models, as long as the model can be sufficiently well approximated on an appropriate time scale as in equation \eqref{eq:3}. Although \cite{Ponulak10} do not explicitly use any neuron model for the derivation of the ReSuMe algorithm, implicitly a linear neuron model is assumed. The algorithm has successfully been applied to leaky integrate-and-fire neurons, Hodgkin-Huxley, and Izhikevich neuron models \citep{Ponulak10}. Since the present learning rule is an extension of ReSuMe to neural networks with multiple layers, this is an indication that this algorithm will function with similar neuron models, as we demonstrate in the following section. 

\subsubsection*{Inhibitory connections}

Inhibitory connections are represented by negative weights which are updated in the same manner as positive weights. However, for the calculation of the backpropagation error of the hidden neurons $\delta_h(t)$ in equation \eqref{eq:22a}, the absolute value of the output weights will be used. This is a deviation from the gradient descent rule, but using the absolute values guarantees that the weights between the input and hidden neurons are always modified in the same direction as between hidden and output neurons:
\begin{equation}
\begin{split}
\Delta w_{hi}(t)= & \frac{1}{n_in_h}S_i(t)\sum_{o\in O}\left[\int_0^\infty a^{pre}[S_o^d(t)-S_o^a(t)]\right]|w_{oh}| \\
                  & + \frac{1}{n_in_h}\sum_{o\in O}\left[S_o^d(t) - S_o^a(t)\right]\left[a+\int_0^\infty a^{post}(s)S_i(t-s)ds\right]|w_{oh}|.
\end{split}
\label{eq:20a}
\end{equation}
Preliminary simulations have shown this results in better convergence of the learning algorithm. There is also neurobiological evidence that LTD and LTP spread to downstream synapses \citep{tao:00, fitzsimonds:97}, i.e. that weight changes with the same direction propagation from upstream to downstream neurons.

\subsection*{Delayed sub-connections}
\label{sec:3.3}
\par
If one considers a network architecture where all the neurons in one layer are connected to all neurons in the subsequent layer through multiple sub-connections with different delays $d_k$, where each sub-connection has a different weight \citep{Bohte02}, the learning rule for the weight modifications for the output neurons will become:
\begin{equation}
\Delta w_{oh}^k=\delta_o(t) R_h(t-d^k_{oh})
\label{eq:21}
\end{equation}
where $w_{oh}^k$ is the weight between output neuron $o$ and hidden neuron $h$ delayed by $d^k_{oh}$ ms.
The backpropagated error for the output is then:
\begin{equation}
\delta_o(t)=\frac{1}{mn_h}\left[R_o^d(t)-R_o^a(t)\right]
\end{equation}
where $m$ is the number of sub-connections. The learning rule for the weight modifications for any hidden layer is derived similarly as:
\begin{equation}
\Delta w_{hi}^k=\delta_h(t) R_i(t-d^k_{oh})
\label{eq:22}
\end{equation}
where $\delta_h(t)$ is the backpropagated error calculated over all possible backward paths (from all output neurons through all delayed sub-connections):
\begin{equation}
\delta_h(t)=\frac{1}{mn_i}\sum_{l, o\in O}\delta_ow_{oh}^l
\label{eq:22a}
\end{equation}

\par
The algorithm can be generalised for neural networks with multiple hidden layers and delays similarly.

\subsection{Synaptic scaling}
\label{sec:3.4}
\par
There has been extensive evidence that suggests that spike-timing dependent plasticity is not the only form of plasticity \citep{Watt10}. Another plasticity mechanism used to stabilise the neurons activity is synaptic scaling \citep{Shepard09}. Synaptic scaling regulates the strength of synapses in order to keep the neuron's firing rate within a particular range. The synaptic weights are scaled multiplicatively, this way maintaining the relative differences in strength between any inputs \citep{Watt10}. 

\par
In our network, in addition to the learning rule described above, the weights are also modified according to synaptic scaling in order to keep the postsynaptic neuron firing rate within an optimal range $[r_{min}, r_{max}]$. If a weight $w_{ij}$ from neuron $i$ to neuron $j$ causes the postsynaptic neuron to fire with a rate outside the optimal range, the weights are scaled according to the following formula \citep{Gruning11}:
\begin{equation}
w_{ij} = \begin{cases}
(1+f)w_{ij}, w_{ij}>0 \\
\frac{1}{1+f}w_{ij}, w_{ij}<0
\end{cases}
\end{equation}
where the scaling factor $f>0$ for $r_j<r_{min}$, and $f<0$ for $r_j>r_{max}$.  

\par
Synaptic scaling solves the problem of optimal weight initialisation. It was observed that the initial values of the weights have a significant influence on the learning process, as too large or too low values may result in failure of the learning \citep{Bohte02}. Preliminary experiments show that a feed-forward network can still learn reliably simple spike trains without synaptic scaling as long as the weights are initialised within an optimal range. However, as the target patterns  contain more spikes, finding the optimal initial values for the weights  becomes difficult. Moreover, as the firing rate of the target neurons increases, it becomes harder to maintain the output neurons firing rate within the target range without using minimal learning steps. The introduction of synaptic scaling solves the problem of weights initialisation as well as speeds up the learning process.

\section{Heuristic discussion of the learning rule}
\label{sec:4}

In order to analyse the direction in which the weights change during the learning process using equations \eqref{eq:12} and \eqref{eq:20a}, we will consider a simple three layer network. The output layer consists of a single neuron. The neurons are connected through a single sub-connection with no delay. For clarity, in this section spike trains will comprise only a single spike. Let $t_d$ and $t_a$ denote the desired and actual spike time of output neuron $o$, and $t_h$ and $t_i$ the spikes times of the hidden neuron $h$ and input neuron $i$ respectively. Also, for simplicity, synaptic scaling will not be considered here.

For a start we assume $t_o, t_d \geq t_h, t_i$, i.e. where relevant post synaptic spikes occur after the pre-synaptic spikes. With these assumptions \eqref{eq:12} and \eqref{eq:20a} read after integrating out: 
\begin{gather}
  \Delta w_{oh}=\frac{1}{n_h}\left(
    A_+\exp{\frac{t_h-t_d}{\tau_+}}
    - A_+\exp{\frac{t_h-t_o}{\tau_+}}\right), \label{eq:31} \\
  \Delta w_{hi}=\frac{1}{n_h n_i}|w_{oh}|\left(
    A_+\exp{\frac{t_i-t_d}{\tau_+}} -
    A_+\exp{\frac{t_i-t_o}{\tau+}}\right). \label{eq:32}
\end{gather}

We discuss this case only in the following and note the case $t_o, t_d < t_h, t_i$ (i.e. post-before-pre) can be discussed along the same lines with $A_+$ above replaced by $A_-$. 

We discussed the following cases:
\begin{enumerate}
\item 
  The output neuron fires a spike at time $t_o$ before the target firing time $t_d$ ($t_o<t_d$). 

  \begin{enumerate}[(a)]
  \item 
    \emph{Weight modifications for the synapses between the output and hidden neurons.} The weights are modified according to $\Delta w_{oh} =\frac{1}{n_h}(A_+\exp{\frac{t_h-t_d}{\tau_+}} -  A_+\exp{\frac{t_h-t_o}{\tau+}})$. Since $t_o<t_d$ then $\exp{(\frac{t_h-t_o}{\tau_+})} > \exp{(\frac{t_h-t_d}{\tau_+})}$ in equation \eqref{eq:31}. This results in $\Delta w_{oh}<0$, and thus in a decrease of this weight. If the connection is an excitatory one, the connection becomes less excitatory, increasing the likelihood, that the output neuron fires later during the next iteration, hence minimising the difference between the actual output and the target firing time. If the connection is inhibitory, the connection will become stronger inhibitory, resulting in a later firing of the output neuron $o$ as well (see also \cite{Ponulak06}).  
  
  \item 
    \emph{Weight modifications for the synapses between the hidden and input neurons.} The weights to the hidden neurons are modified according to: $\Delta w_{hi} =\frac{1}{n_h n_i}(A_+\exp{\frac{t_i-t_d}{\tau_+}} - A_+\exp{\frac{t_i-t_o}{\tau+}})|w_{oh}|$.  
    \par
    \begin{enumerate}[(i)]
    \item $w_{oh} \geq 0$. By an analogue reasoning to the case above $\Delta w_{hi}$, and hence the connection will become less excitatory and more inhibitory, again making the hidden neuron fire a bit later,
and hence making it more likely that also the output neuron fires later as the connection from hidden to output layer is excitatory.
    
 \item 
      $w_{oh} < 0$. For the hidden neuron the effect stays the same, hence it will fire later. As it is now more  likely to fire later, its inhibitory effect will come to bear on the output neurons also a bit later. 
    \end{enumerate}
  \end{enumerate}

\item 
The output neuron fires a spike at time $t_o$ after the target firing time $t_d$ ($t_o>t_d$). As \eqref{eq:31} and \eqref{eq:32} change their sign when $t_o$ and $t_d$ are swapped, this case reduces to the above, but with the opposite sign of the weight change, i.e. overall weight change such that $t_o$ moves forward in time, close to $t_d$.

\end{enumerate}
\par
Cases where there is only an actual spike at $t_o$ and no desired spike or where there is only a desired spike at $t_d$ can be dealt with under the above cases if one sets $t_d = \infty$ or $t_o = \infty$ respectively. In addition there will be a contribution from the factor $a$ in equations \eqref{eq:13} and \eqref{eq:20a}, and this has the same sign as the one from \eqref{eq:31} and \eqref{eq:32}.

\section{Simulations}
\label{sec:5}
In this section several experiments are presented to illustrate the learning capabilities of the algorithm. The algorithm is applied to classic benchmarks, the XOR problem and the Iris data set, as well as to classification tasks with randomly generated patterns. The XOR problem is applied using two different encoding methods to demonstrate the flexibility of our learning algorithm. The learning rule is also applied to classification problems of spike timing patterns which range from 100 ms to 500 ms in order to simulate sensory and motor processing in biological systems.

\subsubsection*{Setup}
\par
The network used for the following simulations is a feed-forward architecture with three layers. The neurons are described by the Spike Response Model \citep{Gerstner01} (see the appendix for a complete description).

\par
For all simulations, an iteration consists of presenting all spike timing pattern pairs in random order. The membrane potential of all neurons in the hidden and output layers is set to the resting potential (set to zero) when presenting a new input pattern. After each presentation of the input pattern to the network, the weight modifications are computed for all layers and then applied. We apply the weight changes after the backpropagated error is computed for all units in the network. The summed network error is calculated for all patterns and tested against a required minimum value, depending on the experiment (see the appendix for details on the network error). This minimum value is chosen in order to guarantee that the network has learnt to correctly classify all the patterns with an acceptable precision. 

\par
The results are averaged over a large number of trials (50 trials unless stated otherwise), with the network being initialised with a new set of random weights every trial. On each testing trial the learning algorithm is applied for a maximum of 2000 iterations or until the network error has reached the minimum value. 

\par
The learning is considered converged if the network error has reached a minimum value, depending on the experiment. Additional constrains for the convergence of the learning algorithm are considered in Sections 5.3 to 5.5 in order to ensure the network has learnt to correctly classify all the patterns. For all simulations, the averaged number of iterations needed for convergence is calculated over the successful trials. The accuracy rate is defined as the percentage of correctly classified patterns calculated over the successful trials.

\par
Unless stated otherwise, the network parameters used in these simulations are: the threshold $\vartheta=0.7$, the time constant of the spike response function $\tau=7$ ms, the time constant of after-potential kernel $\tau_r=12$ ms. The scaling factor is set to $f=\pm 0.005$. The learning parameters are initialised as follows: $A_+ = 1.2$, $A_-=0.5$, $\tau_+=\tau_-=5$ ms, $a=0.05$. 

\par
The weights were initialised with random values uniformly distributed between -0.2 and 0.8. The weights are then normalised by dividing them to the total number of sub-connections.

\subsection{The XOR benchmark}
\label{sec:5.1}
\par
In order to demonstrate and analyse the new learning rule, the algorithm is applied to the XOR problem. While this benchmark does not require generalising, the XOR logic gate is a non-linear problem and it is a classical benchmark for testing the learning algorithm's ability to train non-trivial input output transformations \citep{Rojas96}.
\subsubsection*{Technical details}
\par
The input and output patterns are encoded using spike-time patterns as in \cite{Bohte02}. The signals are associated with single spikes as follows: a binary symbol "0" is associated with a late firing (a spike at 6 ms for the input pattern) and a "1" is associated with an early firing (a spike at 0 ms for the input pattern). We also used a third input neuron that designates the reference start time as this encoding needs an absolute reference start time to determine the latency of the firing \citep{Sporea11}. Without a reference start time, two of the input patterns become identical and without an absolute reference time, the network is unable to distinguish the two patterns (0-0 and 6-6) and would always respond with a delayed output. Table \ref{table1} shows the input and target spike timing patterns that are presented to the network. The values represent the times of the spikes for each input and target neuron in ms of simulated time.

\begin{table}[h!t!]
\caption{Input and output spike-time patterns. The patterns consists of the timing of single spikes in ms of simulated time for the input and target neurons.}
\begin{center}
\begin{tabular}{|l|l|l|l|}
\noalign{\smallskip}
\hline
\multicolumn{3}{|l|} {Input [ms]\rule{0pt}{2.6ex} \rule[-1.2ex]{0pt}{0pt}} & Output [ms]\\
\hline
0 & 0 & 0 & 16\\
\hline
0 & 6 & 0 & 10\\
\hline
6 & 0 & 0 & 10\\
\hline
6 & 6 & 0 & 16\\
\hline
\noalign{\smallskip}
\end{tabular}
\end{center}
\label{table1}
\end{table}

\par
The learning algorithm was applied to a feed-forward network as described above. The input layer is composed of three neurons, the hidden layer contains five spiking neurons, and the output layer contains only one neuron. Multiple sub-connections with different delays were used for each connection in the spiking neural network. Preliminary experiments showed that 12 sub-connections with delays from 0 ms to 11 ms are sufficient to learn the XOR problem. The results are averaged over 100 trials. The network error is summed over all pattern pairs, with a minimum value for convergence of 0.2. The minimum value is chosen to ensure that the network has learnt to classify all patterns correctly, by matching the exact number of spikes of the target spike train as well as the timing of the spikes with 1 ms precision. Each spiking neuron in the network was simulated for a time window of 30 ms, with a time step of 0.1 ms. In the following we systematically vary the parameters of the learning algorithm and examine their effects.

\subsubsection*{The learning parameters}

\par
Here, we vary the learning parameters $A_+$ and $A_-$ in equation \eqref{eq:15w} in order to determine the most appropriate values. $A_+$ is varied between 0.5 and 2.0, while keeping $A_-=\frac{1}{2}A_+$. Table \ref{table23}a shows the summarised results.

\begin{table}[h!t!]
\caption{Summarised results for the XOR problem: (a) The parameters $A_+$ and $A_-$ are varied in order to determine the best values for faster convergence. The ratio between these parameters is constant $A_+=2A_-$. (b) While keeping $A_+=1.2$ fixed, $A_-$ is varied in order to determine the best ratio between these parameters.}
\centering
\subfloat[]{%
\begin{tabular}{|l|l|l|}
\noalign{\smallskip}
\hline
$A_+$ & Successful   & Average number  \\
      &  trials [\%] & of iterations \\
\hline
0.5  & 97 & $331\pm 46$ \\
\hline
0.6  & 98 & $232\pm 24$ \\
\hline
0.7  & 95 & $262\pm 38$ \\
\hline
0.8  & 97 & $144\pm 35$ \\
\hline
0.9  & 96 & $184\pm 23$ \\
\hline
1.0  & 96 & $204\pm 34$ \\
\hline
1.1  & 92 & $166\pm 27$ \\
\hline
1.2  & 96 & $207\pm 31$ \\
\hline
1.3  & 95 & $174\pm 30$ \\
\hline
1.4  & 97 & $183\pm 28$ \\
\hline
1.5  & 93 & $204\pm 36$ \\
\hline
1.6  & 93 & $273\pm 43$ \\
\hline
1.7  & 96 & $163\pm 27$ \\
\hline
1.8  & 94 & $181\pm 32$ \\
\hline
1.9  & 95 & $221\pm 32$ \\
\hline
2.0  & 89 & $141\pm 18$ \\
\hline
\noalign{\smallskip}
\end{tabular}}
\qquad
\subfloat[]{%
\begin{tabular}{|l|l|l|}
\noalign{\smallskip}
\hline
$A_-$ & Successful   & Average number  \\
      &  trials [\%] & of iterations \\
\hline
0.00 & 97 & $231\pm 30$ \\
\hline
0.10 & 98 & $196\pm 24$ \\
\hline
0.20 & 96 & $157\pm 16$ \\
\hline
0.30 & 96 & $187\pm 28$ \\
\hline
0.40 & 95 & $204\pm 37$ \\
\hline
0.50 & 98 & $137\pm 16$ \\
\hline
0.60 & 96 & $207\pm 31$ \\
\hline
0.70 & 95 & $191\pm 33$ \\
\hline
0.80 & 98 & $185\pm 31$ \\
\hline
0.90 & 86 & $203\pm 31$ \\
\hline
1.00 & 88 & $200\pm 30$ \\
\hline
1.10 & 80 & $257\pm 33$ \\
\hline
1.20 & 70 & $349\pm 42$ \\
\hline
1.30 & 65 & $382\pm 30$ \\
\hline
1.40 & 45 & $353\pm 28$ \\
\hline
1.50 & 56 & $492\pm 32$ \\
\hline
\noalign{\smallskip}
\end{tabular}}
\qquad
\label{table23}
\end{table}

\par
The parameters $A_+$ and $A_-$ play the role of a learning rate. Just like the classic back-propagation algorithm for rate neurons, when the learning parameters have higher values the number of iterations needed for convergence is lower. In order to determine the best ratio between the two learning parameters, various values are chosen for $A_-$, while keeping $A_+=1.2$ fixed. The results are summarised in Table \ref{table23}b.

\par
The learning algorithm is able to converge for the values of $A_-$ lower than $A_+$. As $A_-$ becomes equal or higher than $A_+$, the convergence rate slowly decreases and the number of iterations needed for convergence significantly rises. The lowest average number of iterations with a high convergence rate is 137 averaged over 98\% successful trials. 

\subsubsection*{Number of sub-connections}
\par
The algorithm also converges when the spiking neural network has a smaller number of sub-connections. However, a lower number of delayed sub-connections results in a lower convergence rate without necessarily a lower average of learning iterations for the successful trials. Although more sub-connections can produce a more stable learning process, due to the larger number of weights that need to be coordinated, the learning process is slower in this case. Table \ref{table4} shows the summarised results, where $A_+=1.2$ and $A_-=0.6$.

\begin{table}[h!t!]
\caption{The number of delayed sub-connection is varied while keeping the learning parameters fixed $A_+=1.2$ and $A_-=0.6$.}
\begin{center}
\begin{tabular}{|l|l|l|}
\noalign{\smallskip}
\hline
Sub-          & Successful   & Average number  \\
connections   &  trials [\%] & of iterations   \\
\hline
4  & 11  & $63\pm 20$  \\
\hline
6  & 24  & $169\pm 37$ \\
\hline
8  & 73  & $192\pm 27$ \\
\hline
10 & 81  & $154\pm 17$ \\
\hline
12 & 96  & $207\pm 31$ \\
\hline
14 & 96  & $309\pm 52$ \\
\hline
16 & 73  & $472\pm 56$ \\
\hline
\noalign{\smallskip}
\end{tabular}
\end{center} 
\label{table4}
\end{table}

\subsubsection*{Analysis of learning process}
\par
\begin{figure}[h!t!]
\begin{center} 
\leavevmode
\subfloat[]{\includegraphics [height=2.25in] {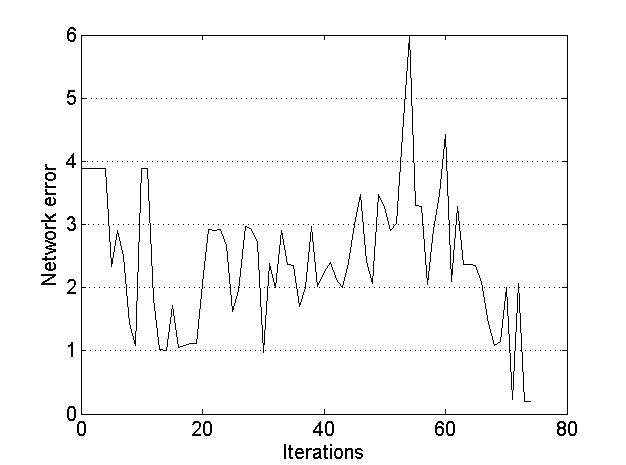} }
\subfloat[]{\includegraphics [height=2.25in] {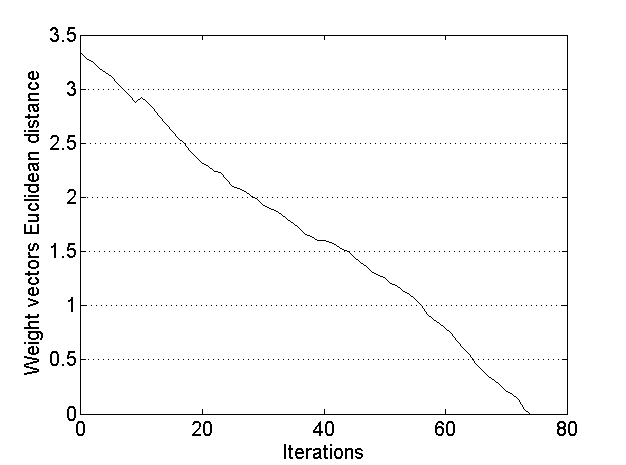} }
\end{center}
\caption{Analysis of the learning process with the parameters $A_+=1.2$ and $A_-=0.5$: (a) The network error during learning. (b) The Euclidean distance between the weight vector solution and the weight vectors during the learning process.}
 \label{fig1}
\end{figure}

In order to analyse the learning process, the network error and the weight vector during the learning process can be seen in Figure \ref{fig1} ($A_+=1.2$, $A_-=0.6$, and 12 sub-connections). Figure \ref{fig1}a shows the evolution of the summed network error during learning. Figure \ref{fig1}b shows the Euclidean distance between the weight vector solution found on a trial and the weight vectors during each learning iteration that led to this weight vector. The weight vectors are tested against the solution found because there can be multiple weight vectors solutions. While the error graph is irregular, the weight vector graph shows that the weight vector moves steadily towards the solution. The irregularity of the network error during the learning process can be explained by the fact that small changes to the weights can produce an additional or missing output spike, which causes significant changes in the network error. The highest error value corresponds to the network not firing any spike for any of the four input patterns. The error graph also shows the learning rule ability to modify the weights in order to produce the correct number of output spikes.

\subsection{The Iris benchmark}
\label{sec:5.2}
\par
Another classic benchmark of pattern recognition is Fisher's Iris flower data set \citep{Fisher36}. The data set contains three classes of Iris flowers. While one of the classes is linearly separable from the other two, the other two classes are not linearly separable from each other. 

\subsubsection*{Technical details}
\par
The three species are completely described by four measurements of the plants: the lengths and wights of the petal and sepal. Each measurement has associated an input neuron and the input pattern consists of the timing of a single spike. The measurements of the Iris flower range from 0 to 8 and are fed into the spiking neural network as spike timing patterns to the input neurons. The output of the network is represented by the spike-time of the output neuron, as seen in Table \ref{table8}. The hidden layer contains ten spiking neurons and each connection has between 8 and 12 delayed sub-connections depending on the experiment. 

\begin{table}[h!t!]
\caption{The target neuron's spike train contains a single spike, where the timing (shown in ms) differs for each of the three patterns.}
\begin{center}
\begin{tabular}{|l|l|}
\noalign{\smallskip}
\hline
Species & Output spike-time [ms] \\
\hline
I. setosa     & 10 \\
\hline
I. versicolor & 14\\
\hline
I. virginica  & 18\\
\hline
\noalign{\smallskip}
\end{tabular}
\end{center}
\label{table8}
\end{table}
\par
During each trial, the input patterns are randomly divided into a training set (75\% of samples) and a testing set (25\% of samples) for cross validation. During each iteration, the training set is used for the learning process to calculate the weight modifications and to test if the network has learnt the patterns. The learning is considered successful if the network error has reach a minimum average value of 0.2 for each pattern pair and 95\% of the patterns in the training set are correctly classified. As in the previous experiment, this minimum value is chosen to ensures that the network has learnt to classify all patterns correctly, by matching the exact number of spikes of the target spike train as well as timing of the spikes with 1 ms precision. Table \ref{table9} shows the summarised results on the Iris data set for different network architectures with different numbers of delayed sub-connections. 

\begin{table}[h!t!]
\caption{Summarised results for the Iris data set.}
\begin{center}
\begin{tabular}{|l|l|l|l|l|}
\noalign{\smallskip}
\hline
Sub-        & Successful & Average number & Accuracy on the   & Accuracy on the  \\
connections & trials     & of iterations  & training set [\%] & testing set [\%] \\
\hline
 8 & 68 & $125\pm 12$ & $97\pm 0.17$ & $89\pm 0.69$ \\
\hline
 9 & 80 & $174\pm 16$ & $96\pm 0.00$ & $94\pm 0.79$ \\
\hline 
10 & 80 & $114\pm 13$ & $97\pm 0.00$ & $89\pm 0.47$ \\
\hline
11 & 74 & $140\pm 15$ & $96\pm 0.16$ & $86\pm 0.49$ \\
\hline
12 & 68 & $183\pm 21$ & $96\pm 0.17$ & $91\pm 0.69$ \\
\hline
\noalign{\smallskip}
\end{tabular}
\end{center}
\label{table9}
\end{table}

\par
Multi-layer ReSuMe permits the spiking neural network to learn the Iris data set using a straight forward encoding of the patterns and results in much faster learning than SpikeProp, as the average number of iterations is always lower than 200, as opposed to the population coding based on arrays of receptive fields that requires 1000 iterations for learning \citep{Bohte02}.

\subsection{Non-linear spike train pattern classification}
\label{sec:5.3}
\par
In this experiment the learning algorithm is tested on non-linear transformation of sequences of spikes. Again, the XOR problem is applied to a network of spiking neurons, but the logic patterns are encoded by spike trains over a group of neurons, and not single spikes (see also \cite{Gruning11}). 

\par
While the encoding for the XOR logic gate problem introduced by \cite{Bohte02} requires neurons to fire a single spike, the network of spiking neurons needs a large number of sub-connections with different delays to enable the hidden and output neurons to fire at the desired times. As the problem becomes more complex such encoding might need even more sub-connections which have to be trained. The large number of weights to be trained slows down the learning process because of the large number of incoming spikes that need to be coordinated to produce the requires output. This can also be seen in the previous simulations on the XOR problem where the network with 14 terminals, although learning the patterns it needed almost twice as many iterations to converge as the network with 12 terminals. Moreover, it has been shown that encoding logical true and false with early and late spike times respectively also requires an additional input neuron to designate the reference start time. Without the additional input neuron, even linear problems become impossible to solve (for a complete demonstration, see \cite{Sporea11}).

\par
A more natural encoding would consist of the temporal firing patterns of groups of neurons \citep{Wehr96, Neuenschwander96, deCharms96}. In order to test such an encoding and the learning algorithm's ability to learn non-linear patterns, the XOR problem is applied once again to a spiking neural network. In this experiment the two logical values will be encoded with spike trains over two groups of input neurons. This encoding will not necessitate multiple delays nor the additional input neuron. In all the following experiments, a single connection with no delay will be used.

\subsubsection*{Technical details}
\par
Each input logical value is associated with the spike trains of a group of 20 spiking neurons. In order to ensure some dissimilarity between the patterns, for each input neuron a spike train is generated by a pseudo Poisson process with a constant firing rate of $r=0.06$ within a 30 ms time window. The minimum inter spike interval is set to 3 ms. This spike train is then split in two new spike trains by randomly distributing all the spikes \citep{Gruning11}. The newly created spike trains will represent the patterns for the logical symbols "0" and "1". The input spike trains are required to consist of at least one spike.

\par
The output patterns are created similarly and will be produced by one output neuron. The spike train to be split is generated by a pseudo Poisson process with a constant firing rate of $r=0.2$ within a 30 ms period of time. The resulting output patterns are chosen so that the spike trains contain exactly three spike.

\par
Apart from the minimal network error as before, an additional stopping criterion for the learning process is introduced. The network must correctly classify all four patterns. An input pattern is considered correctly classified if the output spike train is closest to the target pattern in terms of the van Rossum distance. The network error consist of the sum of van Rossum distances between the target and actual output over the four patterns as before; a minimum value of 3 ensures that the output spikes are reproduced with an acceptable precision.

\par
In addition to the previous experiments, an absolute refractory period is set for all neurons to $t=3$ ms. The learning is simulated over a period of 50 ms, with a time step of 0.5 ms. 

\par
In order to determine the optimal size of the hidden layer for a higher convergence rate, different network topologies have been considered. Table \ref{table5} shows the convergence rate for each network topology, with a new set of spike-timing patterns being generated every trial.

\begin{table}[h!t!]
\caption{Summarised results for the non-linear classifications task.}
\begin{center}
\begin{tabular}{|l|l|l|}
\noalign{\smallskip}
\hline
Hidden  & Successful   & Average number  \\
neurons &  trials [\%] & of iterations \\
\hline
50  & 70  &  $293\pm 59$ \\
\hline
60  & 54  &  $301\pm 66$ \\
\hline
70  & 56  &  $327\pm 91$ \\
\hline
80  & 60  &  $469\pm 87$ \\
\hline
90  & 76  &  $247\pm 42$ \\
\hline
100 & 76  &  $439\pm 73$ \\
\hline
\noalign{\smallskip}
\end{tabular}
\end{center} 
\label{table5}
\end{table}

\par
The learning rule is able to converge with a higher rate as the number of neurons in the hidden layer increases; a larger hidden layer means that the patterns are distributed over a wider spiking activity and easier to be classified by the output neuron. A smaller number of neurons in the hidden layer than in the input layer does not result in high convergence rate because the input patterns are not sufficiently distributed in the hidden activity. Also, more than 100 units in the hidden layer does not result in higher convergence rates, but as the number of weights also increases the learning process is slower. Previous simulations \citep{Gruning11} show that a neural network without a hidden layer cannot learn non-linear logical operations.

\subsection{Learning sequences of temporal patterns}
\label{sec:5.4}
\par
In this experiment, we consider the learning algorithm's ability to train a spiking neural network with multiple input-target pattern pairs. The network is trained with random non-noisy spike train patterns and tested against noisy versions of the temporal patterns.

\subsubsection*{Technical details}
\par
The input patterns are generated by a pseudo Poisson process with a constant firing rate of $r=0.05$ within a 100 ms period of time, where the spike trains are chosen so that they contain at least one spike. In order to ensure that a solution exists, the target patterns are generated as the output of a spiking neural network initialised with a random set of weights. The target spike trains are chosen so they contain at least two spikes and no more than four spikes. If the output patterns were random spike trains, a solution might not be representable in the weight space of the network \citep{Legenstein05}. 

\par
The learning is considered to have converged if the network error reaches an average value of 0.5 for each pattern pair. Apart from the minimum error, the network must also correctly classify at least 90\% of the pattern pairs, where the patterns are classified according the van Rossum distance (see the appendix for details). 

\subsubsection*{The size of the hidden layer}
\par
In order to determine how the structure of the neural network influences the number of patterns that can be learnt, different architectures have been tested. In these simulations, 100 input neurons are considered in order to have a distributed firing activity for the simulated time period. The output layer contains a single neuron as in the previous simulations. The size of the hidden layer is varied from 200 to 300 neurons to determine the optimal size for storing 10 input-output pattern pairs. The results are summarised in Table \ref{table67}a. The network is able to perform better as the number of hidden neurons increases. However, a hidden layer with more 260 neurons does not result in a higher convergence rate.

\begin{table}[h!t!]
\caption{Summarised results for the classification task. (a) The network is trained with 10 pattern pairs, where the size of the hidden layer is varied in order to determine the best network architecture. (b) A neural network with a hidden layer containing 260 neurons is trained with different numbers of pattern pairs.}
\centering
\subfloat[]{%
\begin{tabular}{|l|l|l|}
\noalign{\smallskip}
\hline
Number of & Successful  & Average    \\
hidden    & trials [\%] & number of  \\
units     &             & iterations \\
\hline
200 & 50 & $5\pm 0.8$ \\
\hline
210 & 52 & $6\pm 1.2$ \\
\hline
220 & 78 & $5\pm 0.6$ \\
\hline
230 & 76 & $6\pm 1.1$ \\
\hline
240 & 80 & $5\pm 0.6$ \\
\hline
250 & 74 & $7\pm 0.8$ \\
\hline
260 & 90 & $5\pm 0.7$ \\
\hline
270 & 88 & $4\pm 0.5$ \\
\hline
280 & 80 & $7\pm 2.4$ \\
\hline
290 & 90 & $4\pm 0.6$ \\
\hline
300 & 90 & $4\pm 0.4$ \\
\hline
\noalign{\smallskip}
\end{tabular}
}
\qquad
\subfloat[]{%
\begin{tabular}{|l|l|l|}
\noalign{\smallskip}
\hline
Number of & Successful  & Average    \\
hidden    & trials [\%] & number of  \\
units     &             & iterations \\
\hline
 5 & 100 & $7\pm 0.7$ \\
\hline
 6 &  92 & $5\pm 0.6$ \\
\hline
 7 &  96 & $5\pm 1.2$ \\
\hline
 8 &  92 & $8\pm 1.5$ \\
\hline
 9 &  88 & $7\pm 0.9$ \\
\hline
10 &  90 & $6\pm 0.6$ \\
\hline
11 &  72 & $6\pm 0.7$ \\
\hline
12 &  72 & $6\pm 0.7$ \\
\hline
13 &  58 & $5\pm 0.9$ \\
\hline
14 &  40 & $6\pm 0.9$ \\
\hline
15 &  34 & $5\pm 1.0$ \\
\hline
\noalign{\smallskip}
\end{tabular}
}
\label{table67}
\end{table}

\subsubsection*{Number of patterns}

\par
The networks architecture that performed best with the lowest number of neurons (260 neurons in the hidden layer) was trained with different numbers of patterns. The results for different number of patterns are summarised in Table \ref{table67}b. The network is able to store more patterns, but the convergence rate drops as the number of patterns increases. Because the target patterns are the output spike trains of a randomly initialised spiking neural network, as the number of pattern pairs increases, the target spike trains become necessarily more similar. Hence, the network's responses to the input patterns become more similar and more easily misclassified. Since the stopping criterion requires the network to correctly classify the input patterns, the convergence rate drops as the number of pattern pairs increases.

\par
Since the target patterns are generated as the output spike trains of a network with a set of random weights, this vector of weights can be considered the solution of the learning process. However when looking at the Euclidean distance between the weight vector solution and the weight vectors during learning, the distance is increasing as the learning process progresses. The learning algorithm does not find the same weight vector as the solution, so multiple solutions of weight vectors to the same problem exist (for example permutations of hidden neurons is the simplest one).

\subsubsection*{Noise}
\par
\begin{figure}[h!t!]
\begin{center} 
\leavevmode
\subfloat[]{\includegraphics [height=2.25in] {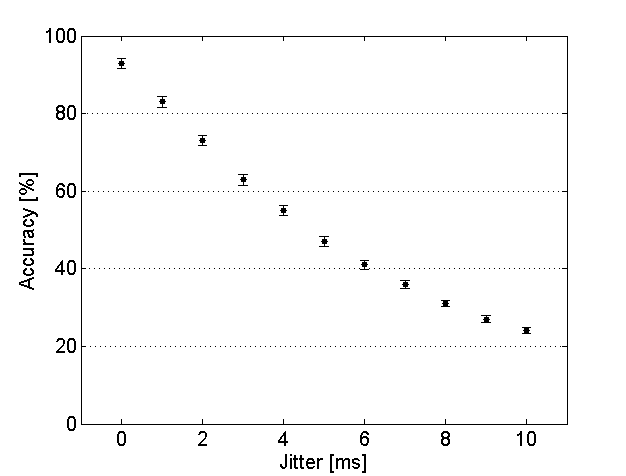} }
\subfloat[]{\includegraphics [height=2.25in] {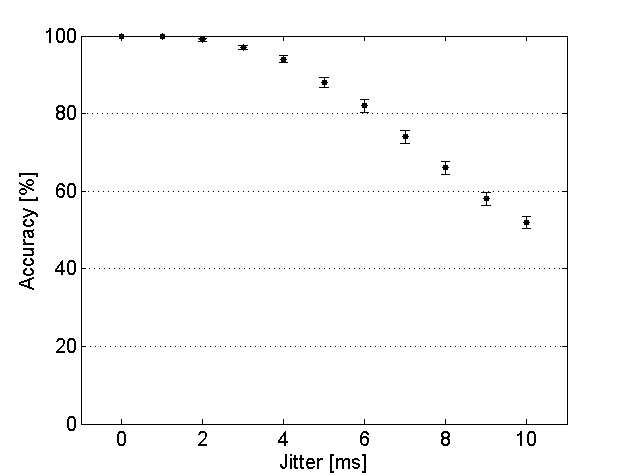} }
\end{center}
\caption{The accuracy on noisy patterns: (a) The network has been trained with 10 non-noisy patterns that span over 100 ms. (b) The network has been trained with 3 noisy patterns that span over 500 ms. During learning the noisy input patterns are generated by moving each spike within a gaussian distribution with mean 0 and standard deviation 4 ms.}
 \label{fig23}
\end{figure}

After the learning has converged, the networks are also tested against noisy patterns. The noisy patterns are generated by moving each spike within a gaussian distribution with mean 0 and standard deviation between 1 and 10 ms. After the network has learnt all patterns, the network is tested with a random set of 500 noisy patterns. Figure \ref{fig23}a shows the accuracy rate (the percentage of input patterns that are correctly classified) for the network with 260 spiking neurons in the hidden layer trained with 10 pattern pairs. The accuracy rates are similar for all the networks described above. The network is able to recognise more than 20\% (above the random performance level of 10\%) of the patterns when these are distorted with 10 ms.

\subsection{Learning to Generalise}
\label{sec:5.5}
\par
In this experiment, the learning algorithm is tested in the presence of noise. In the previous experiments where random patterns were randomly generated, the learning occurred in noise free conditions. A spiking neural network is trained to recognise temporal patterns on the timescale of hundreds of milliseconds. Jitters of spike times are introduced in the temporal patterns during learning to test the network's ability to classify time varying patterns. Such experiments have been conducted with liquid state machines where readout neurons have been trained with ReSuMe to respond with associated spike trains \citep{Ponulak10}. In this paper, we show that such classification tasks can be achieved with feed-forward networks without the need of larger networks such as reservoirs.  

\subsubsection*{Technical details}
\par
Three random patterns are fed into the network through 100 input spiking neurons. The hidden layer contains 210 neurons and the patterns are classified by a single output neuron. The input patterns are generated by a pseudo Poisson process with a constant firing rate of $r=0.1$ within a 500 ms time period, where the spike trains are chosen so that they contain between 15 and 20 spikes. For the spike train generation an inter spike interval is set to 5 ms. As in the previous experiment, in order to ensure that a solution exists, the target patterns are generated as the output of a spiking neural networks initialised with a random set of weights. The target spike trains are chosen so that they contain at least five spikes and no more than seven spikes. The input and target patterns are distributed over such large periods of time in order to simulate complex forms of temporal processing, such as speech recognition, that spans over hundreds of milliseconds \citep{Mauk04}.

\par
During learning, for each iteration noisy versions of the input patterns are generated by moving each spike by a time interval within a gaussian distribution with mean 0 and standard deviation varying in the range of 1 to 4 ms. The spikes in the target patterns are also shifted by a time interval within a gaussian distribution with mean 0 and standard deviation 1 ms independent of the noise level in the input patterns.

\par
A minimum average error of 0.6 for each pattern pair is required for the learning to be considered successful. During each iteration, the network is tested against a new set of 30 random noisy patterns; in order for the learning to be considered converged the network must also correctly classify at least 80\% of noisy patterns. The spike times of the testing patterns are shifted with the same distribution as the training patterns.

\par
Table \ref{table10} shows the convergence rate for each experiment, where the average number of iterations is calculated over the successful trials. The table also shows the number of successful trials when the network is trained on non-noisy patterns. When the network is trained with a low amount of noise in the input patterns, the learning algorithm performs slightly better than the network trained with patterns without noise. The network is able to learn even when the spike train patterns are distorted with 3 or 4 ms. 

\begin{table}[h!t!]
\caption{Summarised results for learning with noisy patterns. The input patterns jitter is varied between 0 and 4 ms, while the output jitter is always 1 ms.}
\begin{center}
\begin{tabular}{|l|l|l|}
\noalign{\smallskip}
\hline
Inputs jitter   & Successful   & Average number  \\
during learning &  trials [\%] & of iterations \\
\hline
0  & 96 & $10\pm 1.2$  \\
\hline
1  & 98 & $12\pm 1.1$  \\
\hline
2  & 95 & $19\pm 2.3$  \\
\hline
3  & 66 & $26\pm 5.6$  \\
\hline
4  & 64 & $115\pm 51$  \\
\hline
\noalign{\smallskip}
\end{tabular}
\end{center}
\label{table10}
\end{table}

\par
Figure \ref{fig23}b shows the accuracy rates on a trained network against a random set of 150 different noisy patterns, generated from the three original input patterns. The network is trained on input patterns where the spikes are moved within a gaussian distribution with mean 0 and variance 4 ms. The graph shows the accuracy rates on patterns with the spikes moved within a gaussian distribution with mean 0 and variance between 1 and 10 ms. The graph also shows the network response on the non-noisy patterns. The accuracy rates are similar for all input pattern jitter. The network is able to recognise more than 50\% (again above the random performance level of 33\%) of the input patterns even when these are distorted with up to 10 ms.

\section{Discussion}
\label{sec:6}
\par
The multilayer ReSuMe permits training spiking neural networks with hidden layers which brings additional computational power. On one hand, the ReSuMe learning rule applied on a single layer \citep{Ponulak10} with 12 to 16 delays for each connection is not able to learn the XOR problem with the early and late timing patterns (see Section \ref{sec:5.1}). Although the algorithm is able to change the weights in the correct direction, the network never responds with the correct output for all four input patterns. The additional hidden layer permits the network to learn the XOR problem (see Section \ref{sec:5.1}). On the other hand, a spiking neural network with the same number of units in each layer, but with 16 sub-connections trained with SpikeProp on the XOR patterns needs 250 iterations to converge \citep{Bohte02}. Our simulations (not presented in this paper) with a similar setup to the experiments in \ref{sec:5.1} confirm this result. Furthermore, SpikeProp requires 16 delayed sub-connections instead of just 12, hence, also implies more weights changes need to be computed. Also, SpikeProp only matches the time of the first target spike, ignoring any subsequent spikes; unlike SpikeProp, our learning algorithm also matches the exact number of output spikes. 

\par
Moreover, studies on SpikeProp show that the algorithm is unstable affecting the performance of the learning process \citep{Takase09, Fujita08}. Our learning algorithm is based on weight modifications that only depend on the timing of pattern pairs and not the specific neuron dynamics, therefore is more stable than SpikeProp (see Figure \ref{fig1}). This can be seen in the direct comparison on the XOR benchmark. Although our algorithm also matches the exact number of spikes as well as the precise timing of the target pattern, the network learns all the patterns faster. 

\par
The learning algorithm presented here permits using different encoding methods with temporal patterns. In section \ref{sec:5.2} the Iris data set is encoded using four input neurons, instead of 50 neurons required by a population encoding \citep{Bohte02}. The simpler encoding of the Iris flower dimensions allows the network to learn the patterns in 5 times less iterations than with a population encoding used with SpikeProp \citep{Bohte02}. 

\par
When moving from rate coded neurons to spiking neurons, an important question about the encoding of patterns arises. One encoding was proposed by \cite{Bohte02}, where logical 0 and 1 are associated with the timing of early and late spikes respectively. As the input neuron's activity is very sparse, the spikes must be multiplied over the simulated time period, as it is known that ReSuMe performs better with more inputs \citep{Ponulak10}. This is achieved by having multiple sub-connections for each input neuron that replicates the action potential with a different delay. The additional sub-connections, each with a different synaptic strength, require additional training. This encoding also requires an additional input neuron to set the reference start time \citep{Sporea11}. Moreover, when looking at the weights after the learning process, only some of the delayed sub-connections have a major contribution to the postsynaptic neuron while others have relatively much smaller absolute values.

\par
The alternative to this encoding is to associate the patterns with spike trains. In order to guarantee that a set of weights exist for any random target transformation without replicating the input signals, a relatively large number of input neurons must be considered. As the input pattern is distributed over several spike trains, some of the information might be redundant and would not have a major contribution to the output. Moreover, such an encoding does not require an additional input neuron to designate the reference start time, as the patterns are encoded in the relative timing of the spikes. The experiment in section \ref{sec:5.3} shows that this encoding can be successfully used for non-linear pattern transformations. 

\par
In the classification task in section \ref{sec:5.4}, where the network is trained on 10 spike-timing pattern pairs, the learning algorithm converges with a higher rate as the hidden layer increases in size. SpikeProp can also be applied to multilayer feed-forward networks but this algorithm is limited to neurons firing a single spike \citep{Bohte02}.

\par
The simulations performed on classification tasks where noise was added to the spike-timing patterns show that the learning is robust to the variability of spike timing. A spiking neural network trained on non-noisy patterns can recognise more than 50\% of noisy patterns if the timing of spikes is shifted with a gaussian distribution with variance up to 4 ms (see Figure \ref{fig23}a), when the network is trained on noisy patterns, it can recognise more than 50\% of noisy patterns where the timing of spikes is moved within a gaussian distributions with variance 10 ms (see Figure \ref{fig23}b).

\par
Another advantage of the learning rule is the introduction of synaptic scaling. Firstly, it solves the problem of finding the optimal range for weight initialisation. This problem is acknowledged as critical for the convergence of the learning \citep{Bohte02}. Secondly, synaptic scaling maintains the firing activity of neurons in the hidden and output layer within an optimal range during the learning process. Although the firing rate of the output and hidden neurons is also adjusted by the non-correlative term $a$ in equations \eqref{eq:12} and \eqref{eq:20a}, this is done only when the output firing rate does not match exactly the target firing rate. This can cause hidden neurons to become quiescent (neurons that do not fire any spike) during the learning process and not to contribute to the activity of the output neurons. Synaptic scaling eliminates this kind of problems by setting a minimum firing rate of one spike.

\section{Conclusion}
\label{sec:7}

\par
This paper introduces a new algorithm for feed-forward spiking neural networks. The first supervised learning algorithm for feed-forward spiking neural networks, SpikeProp, only considers the first spike of each neuron ignoring all subsequent spikes \citep{Bohte02}. An extension of SpikeProp allows multiple spikes in the input and hidden layer, but not in the output layer \citep{Booij05}. Our learning rule is, to the best of our knowledge, the first fully supervised algorithm that considers multiple spikes in all layers of the network. Although ReSuMe allows multiple spikes, the algorithm can only be applied to single layer networks or to train readout neurons in liquid state machines \citep{Ponulak10}. The computational power added by the hidden layer permits the networks to learn non-linear problems and complex classification tasks without using a large number of spiking neurons as liquid state machines do, or without the need of a large number of input neurons in a two layered network. Because the learning rule presented here extends the ReSuMe algorithm to multiple layers, it can in principle be applied to any neuron model, as the weight modification rules only depend on the input, output and target spike trains and does not depend on the specific dynamics of the neuron model.

\section*{Appendix: Details of simulations}
\label{sec:ap}
\subsection*{Neuron model}
\par
The computing units of the feed-forward network used in all simulations are described by the Spike Response Model (SRM) \citep{Gerstner01}. SRM considers the spiking neuron as a homogeneous unit that fires an action potential, or a spike, when the total excitation reaches a certain threshold, $\vartheta$. The neuron is characterised by a single variable, the membrane potential, $u(t)$ at time $t$.

\par
The emission of an action potential can be described by a threshold process as follows. The spike is triggered if the membrane potential $u(t)$ of neuron reaches the threshold $\vartheta$ at time $t^f$:

\begin{equation}
u(t^f)=\vartheta \mbox{ and } \frac{d}{dt}u(t^f)>0 
\end{equation}
 
\par
In the case of a single neuron $j$ receiving input from a set of presynaptic neurons $i\in \Gamma_j$, the state of the neuron is described as follows:
\begin{equation}
u_j(t)=\eta(t-t_j^f)+\sum_{i\in I}\sum_k w_{ji}y_i
\end{equation}
where $y_i$ is the spike response function of the presynaptic neuron $i\in I$, and $w_{ji}$ is the weight between neurons $i$ and $j$; $t_j^f$ is the last firing time of neuron $j$. The kernel $\eta(t)$ includes the form of the action potential as well as the after-potential:

\begin{equation}
\eta(t)=-\vartheta \exp\left(-\frac{t}{\tau_r}\right)
\end{equation}
where $\tau_r>0$ is the membrane time constant, with $\eta(t)=0$ for $t\leq 0$.
\par
The unweighted contribution of a single synaptic to the membrane potential is given by:

\begin{equation}
y_i^k(t)=\sum_{f} \varepsilon\left(t-t_i^f\right)
\end{equation}
with $\varepsilon(t)$ is the spike response function with $\varepsilon(t)=0$ for $t\leq 0$. The times $t_i^f$ represent the firing times of neuron $i$. In our case the spike response function $\varepsilon(t)$ describes a standard post-synaptic potential:

\begin{equation}
\varepsilon(t)=\frac{t}{\tau}\exp\left(1-\frac{t}{\tau}\right)
\end{equation}
where $\tau>0$ models the membrane potential time constant and determines rise and decay of the function.

\subsection*{Network error}
\par
The network error for one pattern is defined in terms of the van Rossum distance between each output spike train and each target spike train \citep{vanRossum01}. The error between the target spike train and the actual spike train is defined as the Euclidean distance of the two filtered spike trains \citep{vanRossum01}. The filtered spike train is determined by an exponential function associated with the spike train:
\begin{equation}
f(t)=\sum_i \exp[-(t-t_i)/\tau_c]H(t-t_i)
\end{equation}
where $t_i$ are the times of the spikes, and $H(t)$ is the Heaviside function. $\tau_c$ is the time constant of the exponential function. $\tau_c$ is chosen to be appropriate to the inter spike interval of the output neurons \citep{vanRossum01}. In the following simulations the output neurons are required to fire approximately one spike in 10 ms, thus $\tau_c=10$ ms. The distance between two spike trains is the squared Euclidean distance between these two functions:
\begin{equation}
D^2(f,g)= \frac{1}{\tau_c}\int^T_0[f(t)-g(t)]^2dt
\end{equation}
where the distance is calculated over a time domain $[0, T]$ that covers all the spikes in the system. The van Rossum distance is also used to determine the output pattern during learning and testing. The output pattern is determined as the closest to one of the target patterns in terms of the van Rossum distance.


\begin{thebibliography}{100}
\providecommand{\natexlab}[1]{#1}
\expandafter\ifx\csname urlstyle\endcsname\relax
  \providecommand{\doi}[1]{doi:\discretionary{}{}{}#1}\else
  \providecommand{\doi}{doi:\discretionary{}{}{}\begingroup
  \urlstyle{rm}\Url}\fi


\bibitem[{Bohte et~al.(2002) Bohte, Kok, \& Poutr\'e}]{Bohte02}
Bohte, S., Kok, J., \& Poutr\'e, H.L. (2002).
\newblock Error backpropagation in temporally encoded networks of spiking neurons.
\newblock \emph{Neurocomputing}, \emph{48}, 17 -- 37.

\bibitem[{Booij \& tat Nguyen(2005)}]{Booij05}
Booij, O. \& tat Nguyen, H. (2005). 
\newblock A gradient descent rule for spiking neurons emitting multiple spikes.
\newblock \emph{Information Processing Letters}, \emph{95(6)}, 552 -- 558.

\bibitem[{deCharms \& Merzenich(1996)}]{deCharms96}
deCharms, R.C. \& Merzenich, M.M. (1996).
\newblock Primary cortical representation of sounds by the coordination of action-potential timing.
\newblock \emph{Nature}, \emph{381}, 610 -– 613.

\bibitem[{Elias \& Northmore(2002)}]{Elias02}
Elias, J.G. \& Northmore, D.P.M. (2002). 
\newblock Building silicon nervous systems with dendritic tree neuromorphs.
\newblock In Maass, W., Bishop, C.M. (Eds.), \emph{Pulsed Neural Networks}, MIT Press, Cambridge.

\bibitem[{Fisher(1936)}]{Fisher36}
Fisher, R.A. (1936).
\newblock The Use of Multiple Measurements in Taxonomic Problems.
\newblock \emph{Annals of Eugenics}, \emph{7(2)}, 179 -– 188.

\bibitem[{Fitzsimonds et~al.(1997)}]{fitzsimonds:97}
Fitzsimonds, R.M., Sonf, H. \& Poo, M. (1997).
\newblock Propagation of activity dependent synaptic depression in simple neural networks.
\newblock \emph{Nature}, \emph{388}, 439 -- 448.

\bibitem[{Fujita et~al.(2008)}]{Fujita08}
Fujita, M., Takase, H., Kita, H., \& Hayashi, T. (2008). 
\newblock Shape of error surfaces in SpikeProp. 
\newblock \emph{Proceedings of IEEE International Joint Conference Neural Networks, IJCNN08}, 840 -- 844.

\bibitem[{Gerstner(2001)}]{Gerstner01}
Gerstner, W. (2001).
\newblock A Framework for Spiking Neuron Models: The Spike Response Model.
\newblock In: Moss, F., Gielen, S. (Eds.), \emph{The Handbook of Biological Physics}, \emph{Vol.4 (12)}, 469 -- 516.

\bibitem[{Gerstner \& Kistler(2002)}]{Gerstner02}
Gerstner, W. \& Kistler, W.M. (2002).
\newblock \emph{Spiking Neuron Models. Single Neurons, Populations, Plasticity}, Cambridge University Press, Cambridge.


\bibitem[{Glackin et~al.(2011)}]{glackin:11}
Glackin, C., Maguire, L., McDaid, L. \& Sayers, H. (2011).
\newblock Respective field optimisation and supervision of a fuzzy spiking neural network.
\newblock \emph{Neural Networks}, \emph{24}, 247 -- 256.

\bibitem[{Gr\"uning \& Sporea(2011)}]{Gruning11}
Gr\"uning, A. \& Sporea, I. (2011).
\newblock Supervised Learning of Logical Operations in Layered Spiking Neural Networks with Spike Train Encoding.
\newblock Submitted for publication.
\newblock Preprint available online at: http://arxiv.org/abs/1112.0213.

\bibitem[{G\"utig et~al.(2003)G\"utig, }]{Gutig03}
G\"utig, R.,  Aharonov, R., Rotter, S., \& Sompolinsky, H. (2003)
\newblock Learning Input Correlations through Nonlinear Temporally Asymmetric Hebbian Plasticity.
\newblock \emph{The Journal of Neuroscience}, \emph{23(9)}, 3697 -- 3714.

\bibitem[{G\"utig \& Sompolinsky(2006)}]{Gutig06}
G\"utig, R. \& Sompolinsky, H. (2006).
\newblock The tempotron: a neuron that learns spike timing-based decisions.
\newblock \emph{Nature Neuroscience}, \emph{9(3)}, 420 -- 428.

\bibitem[Harris(2008)]{harris:08}
Harris, K.D. (2008).
\newblock Stability of the fittest: organizing learning through retroaxonal signals.
\newblock \emph{Trends in Neuroscience}, \emph{31(3)}, 130 -- 136.

\bibitem[{Hebb(1949)}]{Hebb49}
Hebb, D.O. (1949).
\newblock \emph{The organization of behavior}, Wiley, New York.

\bibitem[{Heeger(2001)}]{Heeger01}
Heeger, D. (2001).
\newblock Poisson Model of Spike Generation.
\newblock Available online at: www.cns.nyu.edu/~david/handouts/poisson.pdf.

\bibitem[{Johansson \& Birznieks(2004)}]{Johansson04}
Johansson, R.S. \& Birznieks, I. (2004).
\newblock First spikes in ensembles of human tactile afferents code complex spatial fingertip events.
\newblock \emph{Nature Neuroscience}, \emph{7}, 170 -- 177.

\bibitem[{Legenstein et~al.(2005)Legenstein, Naeger, \& Maass}]{Legenstein05}
Legenstein, R., Naeger, C., \& Maass, W. (2005).
\newblock What can a neuron learn with spike-timing-dependent plasticity? 
\newblock \emph{Neural Computation}, \emph{17(11)}, 2337 -- 2382.

\bibitem[{Kempter et~al.(2001)Kempter, Gerstner, \& Van Hemmen}]{Kempter01}
Kempter, R. , Gerstner, W., \& Van Hemmen, J.L. (2001).
\newblock Intrinsic Stabilization of Output Rates by Spike-Based Hebbian Learning.
\newblock \emph{Neural Computation}, \emph{13}, 2709 -- 2741.

\bibitem[{Knudsen(1994)}]{Knudsen94}
Knudsen, E.I. (1994). 
\newblock Supervised learning in the brain.
\newblock \emph{Journal of Neuroscience}, \emph{14(7)}, 3985 -- 3997.

\bibitem[{Knudsen(2002)}]{Knudsen02}
Knudsen, E.I. (2002). 
\newblock Instructed learning in the auditory localization pathway of the barn owl.
\newblock \emph{Nature}, \emph{417(6886)}, 322 -- 328.


\bibitem[{Maass(1997a)}]{Maass97a}
Maass, W. (1997).
\newblock Networks of spiking neurons: the third generation of neural network models.
\newblock \emph{Transactions of the Society for Computer Simulation International}, emph{Vol. 14 (4)}, 1659 -- 1671.

\bibitem[Maass(1997b)]{Maass97b}
Maass, W. (1997).
\newblock Fast sigmoidal networks via spiking neurons.
\newblock \emph{Neural Computation}, \emph{9}, 279 -- 304.

\bibitem[{Mauk \& Buonomano(2004)}]{Mauk04}
Mauk, M.D. \& Buonomano, D.V. (2004).
\newblock The Neural Basis of Temporal Processing.
\newblock \emph{Annual Rev. Neuroscience}, \emph{27}, 304 -- 340.

\bibitem[{Neuenschwander \& Singer(1996)}]{Neuenschwander96}
Neuenschwander, S. \& W. Singer (1996).
\newblock Long-range synchronization of oscillatory light responses in the cat retina and lateral geniculate nucleus.
\newblock \emph{Nature}, \emph{379}, 728 -- 733.

\bibitem[{Ponulak(2006)}]{Ponulak06}
Ponulak, F. (2006).
\newblock ReSuMe - Proof of convergence.
\newblock Available online at: http://d1.cie.put.poznan.pl/dav/fp/FP\_ConvergenceProof\_TechRep.pdf.

\bibitem[{Ponulak \& Kasi\'{n}ski(2010)}]{Ponulak10}
Ponulak, F. \& Kasi\'{n}ski, A. (2010). 
\newblock Supervised learning in spiking neural networks with ReSuMe: Sequence learning, classification, and spike shifting. 
\newblock \emph{Neural Computation}, \emph{22(2)}, 467 -- 510.

\bibitem[{Rojas(1996)}]{Rojas96}
Rojas, R. (1996).
\newblock \emph{Neural Networks - A Systematic Introduction}, Springer-Verlag, Berlin.

\bibitem[{van Rossum(2001)}]{vanRossum01}
van Rossum, M.C. (2001).
\newblock A novel spike distance.
\newblock \emph{Neural Computation}, \emph{13(4)}, 751 -- 63.

\bibitem[{Rostro-Gonzalez et~al.(2010)}]{rostro:10}
Rostro-Gonzalez, H., Juan Carlos Vasquez-Betancour, J.C., Cessac, B. \& Vi\'eville, T. (2010).
\newblock Reverse-engineering in spiking neural network parameters: exact determinisitc parameter estimation.
\newblock INRIA Sophia Antipolis.

\bibitem[{Ruf \& Schmitt(1997)}]{Ruf97}
Ruf, B. \& Schmitt, M. (1997).
\newblock Learning temporally encoded patterns in networks of spiking neurons. 
\newblock \emph{Neural Processing Letters}, \emph{5(1)}, 9 -- 18.

\bibitem[{Rumelhart et~al.(1986)Rumelhart, Hinton, \& Williams}]{Rumelhart86}
Rumelhart, D.E., Hinton, G.E., \& Williams, R.J. (1986).
\newblock Learning internal representations by error propagation.
\newblock In Rumelhart, D.E. and McClelland, J.L. (Eds.), \emph{Parallel distributed processing: Explorations in the microstructure of cognition}, \emph{Vol. 1}, MIT Press, Cambridge, MA.

\bibitem[{Schrauwen \& Van Campenhout(2004)}]{Schrauwen04}
Schrauwen, B. \& Van Campenhout, J. (2004). 
\newblock Improving Spike-Prop: Enhancements to an Error-Backpropagation Rule for Spiking Neural Networks. 
\newblock \emph{Proceedings of the 15th ProRISC Workshop}.

\bibitem[{Shepard et~al.(2009)Shepard, Rumbaugh, Wu, Chowdhiry, Plath, Kuhl, Huganir, Worley}]{Shepard09}
Shepard, J.D., Rumbaugh, G., Wu, J., Chowdhiry, S., Plath, N., Kuhl, D., Huganir, R.L., \& Worley, P.F.
(2009).
\newblock Arc mediates homoestatoc synaptic scaling of ampa receptors.
\newblock \emph{Neuron}, \emph{52(3)}, 475 -- 484.

\bibitem[{Sporea \& Gr\"uning(2011)}]{Sporea11}
Sporea, I. \& Gr\"uning, A. (2011).
\newblock Reference Time in SpikeProp
\newblock \emph{Proceedings of IEEE International Joint Conference Neural Networks, IJCNN11}, 1090 -- 1092.

\bibitem[{Takase et~al.(2009)}]{Takase09}
Takase, H., Fujita, M., Kawanaka, H., Tsuruoka, S., Kita, H., \& Hayashi, T. (2009).
\newblock Obstacle to training SpikeProp Networks - Cause of surges in training process.
\newblock \emph{Proceedings of IEEE International Joint Conference Neural Networks, IJCNN09}, 3062 -- 3066.

\bibitem[{Ti\v{n}o \& Mills(2005)}]{Tino05}
Ti\v{n}o, P. \& Mills A.J. (2005).
\newblock Learning beyond finite memory in recurrent networks of spiking neurons.
\newblock In Wang, L., Chen, K., Ong, Y. (Eds.), \emph{Advances in Natural Computation - ICNC 2005, Lecture Notes in Computer Science}, 666 -- 675.

\bibitem[{Tao et~al.(2000)}]{tao:00}
Tao, H.W., Zhang, L.I, Bi, G.Q. \& Poo, M. (2000).
\newblock Selective Presynaptic Propagation of Long-Term Potentiation in Defined Neural Networks.
\newblock \emph{Journal of Neuroscience}, \emph{20(9)}, 3233 -- 3243.

\bibitem[{Thorpe \& Imbert(1989)}]{Thorpe89}
Thorpe, S.T. \& Imbert, M. (1989). 
\newblock Biological constraints on connectionist modelling.
\newblock In Pfeifer, R., Schreter, Z., Fogelman-Soulié, F., Steels, L. (Eds.), \emph{Connectionism in perspective},  63 -- 92.

\bibitem[Wade et~al.(2010)]{wade:10}
Wade, J.J., McDaid, L.J., Santos, J.A. \& Sayers, H.M. (2010).
\newblock SWAT: A Spiking Neural Network Training Algorithm for Classification Problems.
\newblock \emph{IEEE Transactions on Neural Networks}, \emph{21(11)}, 1817 -- 1829.


\bibitem[{Watt \& Desai(2010)}]{Watt10}
Watt, A.J. \& Desai, N.S. (2010).
\newblock Homeostatic plasticity and STDP: keeping a neuron's cool in a fluctuating world.
\newblock \emph{Frontiers in Synaptic Neuroscience}, \emph{2(5)}.

\bibitem[{Wehr \& Laurent(1996)}]{Wehr96}
Wehr, M. and Laurent, G. (1996).
\newblock Odour encoding by temporal sequences of firing in oscillating neural assemblies.
\newblock \emph{Nature}, \emph{384}, 162 -- 166.

\bibitem[{Xin \& Embrechts(2001)}]{Xin01}
Xin, J. \& Embrechts, M.J. (2001).
\newblock Supervised Learning with Spiking Neuron Networks. 
\newblock \emph{Proceedings of IEEE International Joint Conference Neural Networks, IJCNN01}, 1772 -- 1777.

\end{thebibliography}
\end{document}